\documentclass[10pt,twocolumn,letterpaper]{article}
\usepackage[pagenumbers]{wacv}

\usepackage{graphicx}
\usepackage{amsmath}
\usepackage{amssymb}
\usepackage{algorithm}
\usepackage[noend]{algpseudocode}
\usepackage{tabularx}
\usepackage{multirow}
\usepackage{booktabs}
\usepackage{array}
\usepackage{bbm}
\usepackage{varwidth}
\usepackage[table,xcdraw]{xcolor}
\usepackage[pagebackref,breaklinks,colorlinks]{hyperref}
\usepackage[capitalize]{cleveref}
\crefname{section}{Sec.}{Secs.}
\Crefname{section}{Section}{Sections}
\Crefname{table}{Table}{Tables}
\crefname{table}{Tab.}{Tabs.}

\renewcommand{\algorithmicrequire}{\textbf{Input:}}
\renewcommand{\algorithmicensure}{\textbf{Output:}}

\begin{document}

\title{\textsc{DragText}: Rethinking Text Embedding in Point-based Image Editing}

\author{
Gayoon Choi\quad 
Taejin Jeong\quad 
Sujung Hong\quad 
Seong Jae Hwang\thanks{Corresponding author}\\ 
Yonsei University
\\{\tt\small \{gynchoi17, starforest, sujung0914, seongjae\}@yonsei.ac.kr}
}

\twocolumn[{%
\renewcommand\twocolumn[1][]{#1}
\maketitle
\vspace{-7pt}
\begin{center}
    \centering
    \captionsetup{type=figure}
    \includegraphics[width=\textwidth]{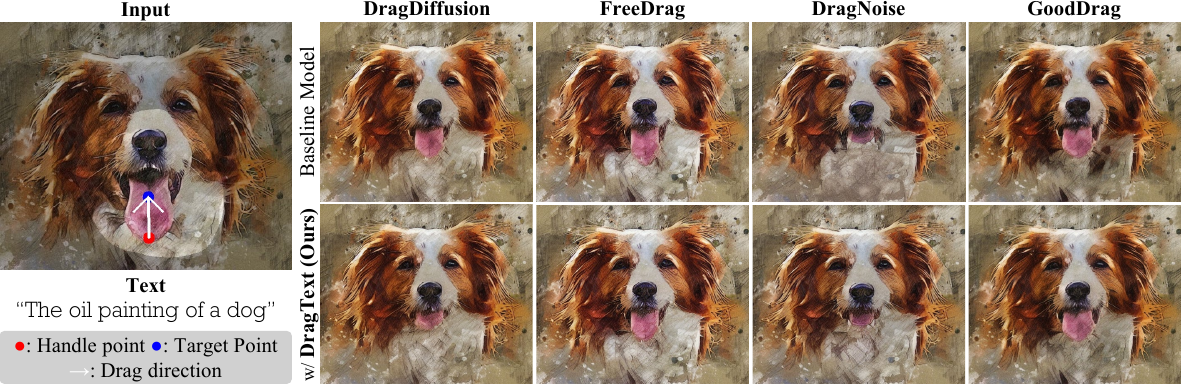}
    \vspace{-18pt}
    \captionof{figure}{In point-based image editing, a user first draws a mask on the image to define an editable region, then edits the image by ``dragging'' the contents from the user-defined handle points (\textcolor{red}{red}) to target points (\textcolor{blue}{blue}).
    Our \textsc{DragText} stabilizes the process of point-based image editing through parallel image and text optimization, consistently improving an array of existing diffusion-based drag methods.}
    \label{fig:teaser}
\end{center}
}]

\let\thefootnote\relax\footnote{
\hspace{-1em}
\begin{minipage}[t]{\textwidth}
\noindent\textsuperscript{*}Corresponding author\\
\noindent\hspace*{0.41em}\scriptsize Project Page:~\url{https://micv-yonsei.github.io/dragtext2025/}
\end{minipage}
}

\vspace{-10pt}
\begin{abstract}\label{sec:abs}
\vspace{-8pt}
Point-based image editing enables accurate and flexible control through content dragging.
However, the role of text embedding during the editing process has not been thoroughly investigated. 
A significant aspect that remains unexplored is the interaction between text and image embeddings. 
During the progressive editing in a diffusion model, the text embedding remains constant. 
As the image embedding increasingly diverges from its initial state, the discrepancy between the image and text embeddings presents a significant challenge. 
In this study, we found that the text prompt significantly influences the dragging process, particularly in maintaining content integrity and achieving the desired manipulation. 
Upon these insights, we propose \textsc{DragText}, which optimizes text embedding in conjunction with the dragging process to pair with the modified image embedding. 
Simultaneously, we regularize the text optimization process to preserve the integrity of the original text prompt. 
Our approach can be seamlessly integrated with existing diffusion-based drag methods, enhancing performance with only a few lines of code.
\vspace{-20pt}
\end{abstract}

\vspace{-30pt}
\section{Introduction}\label{sec:intro}
\vspace{6pt}

Text-to-image (T2I) models have made significant advancements in image editing alongside the development of diffusion models~\cite{Dalle1, StableDiffusion, Dalle2, Imagen, Kandinsky, Pixart}.
These models are effective in broad modifications, such as inpainting\cite{InpaintAnything, Smartbrush}, style transfer~\cite{TI, InST}, and content replacement~\cite{dreambooth, prompt, Imagic}.
For instance, when a user inputs the text prompt ``\texttt{a dog looking up}'' into a T2I model, the resulting image shows the dog lifting its head.
However, if one wishes to provide clearer explicit instructions for more detailed structural edits (e.g., movement angle or distance), designing appropriate text prompts to inject such a level of intention becomes far from trivial.
To deal with this, several studies have shifted towards non-text controls (\eg, points, edges, poses, sketches) to avoid ambiguity and achieve controllability~\cite{DragGAN, ControlNet, PromptFree, T2iAdapter}. 
Among these, \textit{point-based image editing} (\cref{fig:teaser}) is particularly noteworthy in that it employs pairs of instruction points, allowing fine-grained control.

Point-based image editing has recently advanced with diffusion models, effectively manipulating both synthetic and real images across diverse domains~\cite{DragGAN, DragDiffusion, FreeDrag, DragNoise, GoodDrag}.
To generalize image editing in various fields, large-scale pre-trained latent diffusion models have been utilized.
These models are trained with cross-attention between text and image embeddings in denoising U-Net.
Therefore, they also require text prompts as input. 
When point-based image editing drags content in the image embedding space, the text prompt provides conditions to the latent vector.
However, to the best of our knowledge, no prior research has investigated the impact of text.
Unlike T2I methods which actively analyze text and utilize text in innovative ways~\cite{TI, NTI, prompt, CXRL, cat}, the role of text and its potential in point-based image editing remains unknown.

\begin{figure}[htb!]
    \centering
    \includegraphics[width=0.47\textwidth]{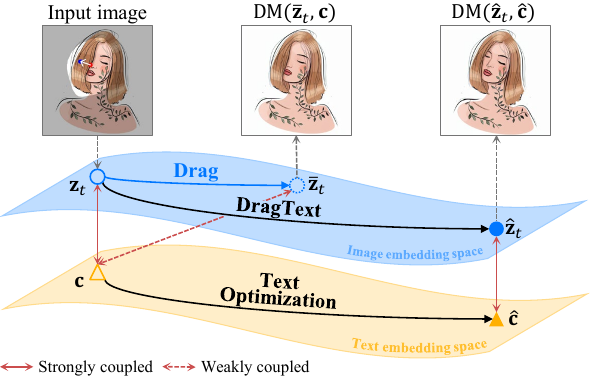}
    \vspace{-5pt}
    \caption{Illustration of the drag editing process within the image and text embedding spaces of the diffusion model (DM). 
    During editing, the original image embedding $\mathbf{z}_t$ naturally deviates to the dragged image latent vector $\mathbf{\bar{z}}_t$. 
    Without text optimization, the corresponding text embedding $\mathbf{c}$ is decoupled from $\mathbf{\bar{z}}_t$, resulting in drag halting.
    Hence, optimal text embedding $\mathbf{\hat{c}}$ coupled with dragged images has to be acquired to make the optimal latent vector $\mathbf{\hat{z}}_t$ which then holds the related semantics via text.}
    \vspace{-.4cm}
    \label{fig:insight}
\end{figure}

In this study, we explore how the text prompt affects the dragging process and discover whether the content successfully dragged to the desired position is influenced by the text prompt. 
As the image is optimized through drag editing, it naturally deviates from the original image in the image embedding space~\cite{SDEDrag}.
However, the text embedding remains stationary and thus fails to describe the edited image accurately.
This static text embedding is used during the drag editing process and the denoising process in point-based image editing~\cite{DragDiffusion, FreeDrag, DragNoise, GoodDrag}, leading to what we term \textit{drag halting}, where the drag editing process fails to reach the intended drag points, or it reaches them but with a loss of semantic integrity.
For example, as shown in \cref{fig:insight}, during the editing process of moving a woman's hair aside, the edited image is no longer strongly coupled with the original text ``\texttt{A woman with a leaf tattoo on her neck.}''.
Despite the change in the woman's hair, the original text prompt does not reflect this change.
As a result, the handle point falls short of fully reaching the target points as seen in the middle image. 

We propose \textsc{DragText}, a method designed to rectify the static text embedding in point-based image editing.
\textsc{DragText} optimizes the original text embedding in a similar way to image optimization.
``Dragging'' text embedding in parallel with the image ensures that the edited text embedding remains coupled with the edited image embedding.
It alleviates drag halting by optimizing the image embedding, which helps handle points reach their targets.
Moreover, we designed a text embedding regularization technique to ensure the edited text embedding does not diverge too significantly from the original text embedding.
It preserves key styles of the image while allowing for natural changes within the image during drag editing.

\noindent\textbf{Contributions.} Our contributions are as follows: 
1) We are the first to analyze the impact of text prompts on the dragging process, a previously neglected yet crucial component of point-based image editing; 
2) We propose \textsc{DragText} which optimizes text embedding that aids drag operations without forgetting content; 
3) \textsc{DragText} can be easily applied in a plug-and-play manner to various diffusion-based point-based image editing methods, consistently improving performance.
\section{Related Work}\label{sec:rw}

\subsection{Point-based Image Editing}
\noindent\textbf{Task Definition.} 
Point-based image editing enables precise modifications to images via point instruction~\cite{UserControllableLT, DragGAN, DragonDiffusion, DiffEditor}.
As shown in \cref{fig:teaser}, a user inputs the image to edit, draws a mask representing an editable region, and specifies handle points along with their corresponding target locations.
During the editing process, handle points are moved toward their respective target points through an image optimization strategy.
The positions of these points are continuously tracked and updated after each optimization step.

\vspace{3pt}
\noindent\textbf{Existing Methods.} 
DragGAN~\cite{DragGAN} is a representative method in point-based image editing that proposes latent code optimization and point tracking.
However, it struggles with generalization, particularly with real images.
DragDiffusion~\cite{DragDiffusion} employs a text-conditional diffusion model~\cite{StableDiffusion} to expand applicability and improve spatial control, following the strategy of DragGAN.
Several point-based editing diffusion methods have been developed upon these advancements. 
For instance, FreeDrag~\cite{FreeDrag} uses original image features as a reference template and reduces the complexity of point tracking with a line search.
DragNoise~\cite{DragNoise} reduces computational load by optimizing the U-Net bottleneck feature instead of the latent vector. 
GoodDrag~\cite{GoodDrag} alternates between dragging and denoising processes to prevent error accumulation and enhance image fidelity.
Previous works have primarily focused on image optimization to directly improve manipulation. 
Unlike these, our approach emphasizes the underestimated role of the text prompt, providing a new perspective of enhancing not only manipulation quality but also the dragging mechanism.

\subsection{Text Optimization}
Recent advancements in Vision-Language Models (VLMs) have significantly enhanced the ability to flexibly connect text and images~\cite{CLIP, ALIGN}. 
Various methods of text optimization are being explored to improve these connections. 
One approach is context optimization~\cite{CoOp, CoCoOp}, which refines text prompts to help VLMs better understand hand-crafted captions.
Another method, developed by Gal \etal~\cite{TI}, optimizes a single word to effectively convey content information, thereby personalizing text-based image editing. 
Mokady \etal~\cite{NTI} introduces the strategy to optimize null-text ``\texttt{}'', which addresses the issue of images deviating from their original trajectory due to accumulated errors from classifier-free guidance~\cite{CFG}. 
Our work is inspired by these prior studies which closely examine the strong coupling between text and image.
However, our approach differs from these in that we perform text optimization simultaneously with image optimization in point-based image editing to avoid increasing the optimization steps and time.
\section{Motivations}\label{sec:motiv}
Editing an image causes the original image embedding to move in the image embedding space.
Therefore, the edited image embedding must follow the denoising trajectory that differs from that of the original image embedding~\cite{SDEDrag, NTI}. 
We hypothesize that text embedding also plays a crucial role in determining this distinct trajectory as text and image interact through cross-attention.

\subsection{Text Prompt in Diffusion Model}\label{subsec:motiv1}
\noindent\textbf{Role in Inversion and Denoising Processes.} 
\textit{The inversion process pairs the image embedding with text embedding, regardless of whether the text embedding semantically matches the image.
Maintaining this pairing in the denoising process is important for the fidelity of image sampling.}

\begin{figure}[htb!]
    \centering
    \includegraphics[width=0.47\textwidth]{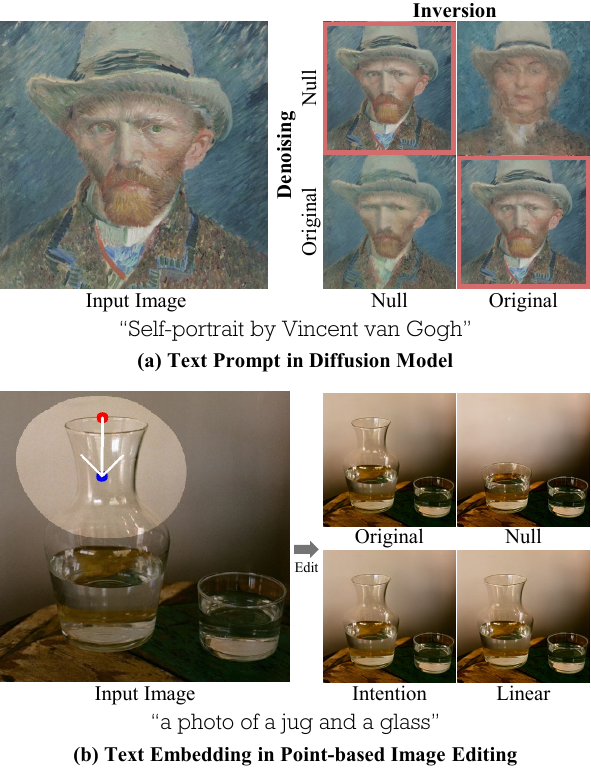}
    \vspace{-5pt}
    \caption{(a) Red boxes imply the consistent use of text prompts across diffusion processes, resulting in high-fidelity sampling.
    In contrast, other images result from inconsistent text prompt usage, leading to inaccurate sampling outcomes.
    (b) The original text prompt and its alternative intention text prompt, crafted from prompt engineering, are insufficient to prevent drag halting.}
    \vspace{-10pt}
    \label{fig:motiv}
\end{figure}

In \cref{fig:motiv} (a), we examine differences in image sampling outcomes based on whether the text embedding is also paired in the denoising process.
Our analysis reveals that using the paired text embedding in the denoising process (red boxes) enables the accurate sampling of the original image.
However, if unpaired text embedding is employed, the image's style alters, and the model fails to sample the image accurately.
This observation highlights the necessity of text-image embedding pairing to maintain the integrity of the image's attributes.
Moreover, it raises concerns about the appropriateness of using the original text embedding in the denoising process of point-based image editing, as the image is modified during dragging, it can no longer be accurately paired with the original text embedding.

\vspace{3pt}
\noindent\textit{\textbf{Remark.}} To achieve the appropriate image sampling result, the text embedding used after the inversion process remains consistently paired with the image embedding.

\begin{figure*}[t]
    \centering
    \includegraphics[width=\textwidth]{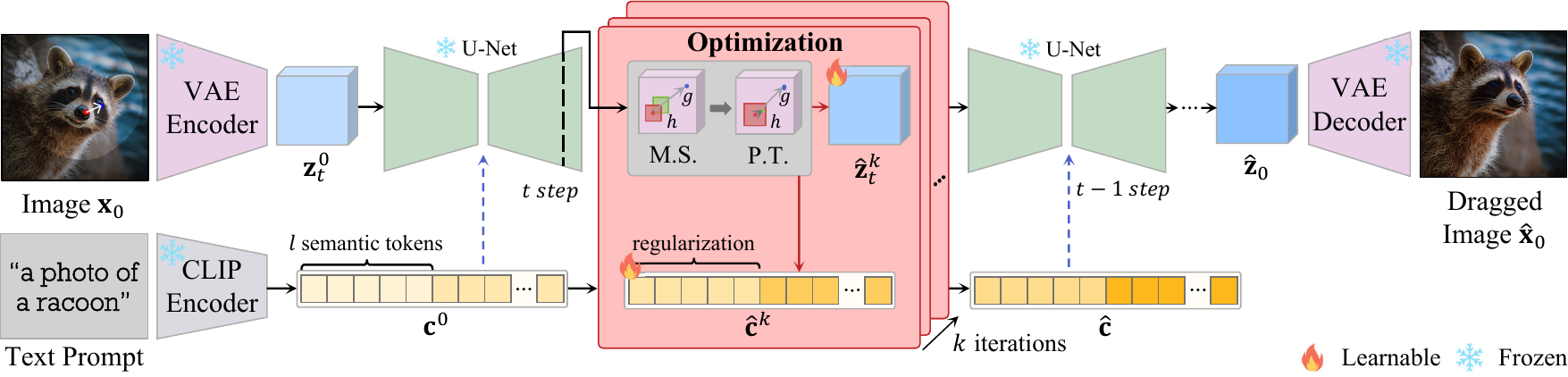}
    \vspace{-15pt}
    \caption{The pipeline of \textsc{DragText}.
    The image $\mathbf{x}_0$ is encoded through a VAE encoder as the latent vector $\mathbf{z}$, and the text is encoded by a CLIP text encoder as the text embedding $\mathbf{c}$.
    Through DDIM inversion with $\mathbf{c}$, the latent vector $\mathbf{z}_t$ is obtained.
    At time step $t=35$, $\mathbf{z}^0_t$ and $\mathbf{c}$ are optimized to $\mathbf{\hat{z}}^k_t$ and $\mathbf{\hat{c}}$ by iterating with motion supervision (M.S.), text optimization, and point tracking (P.T.) $k$-times.}
    \label{fig:pipeline}
    \vspace{-10pt}
\end{figure*}

\subsection{Text Embedding in Point-based Image Editing}\label{subsec:motiv2} 
\noindent\textbf{Challenges in Original Text.} 
\textit{During dragging, the image embedding incrementally adapts to reflect the manipulation.
However, the text embedding remains static, limiting the extent of image manipulation.}

As shown in the top-left image of \cref{fig:motiv} (b), the original text embedding fails to reach the target point during the dragging process.
Since the original text embedding is paired with the input image, the edited image embedding does not maintain a strong coupling with the text embedding. 
To avoid this weakly-coupling issue, one might hypothesize that unconditional text embedding (\ie null text) could resolve the problem.
However, the top-right image of \cref{fig:motiv} (b) shows that the model fails to maintain the semantic information of the object without text conditions.
Therefore, we search for an alternative text embedding that mitigates drag halting while preserving semantics.

\vspace{3pt}
\noindent\textit{\textbf{Remark.}} While text embedding plays a crucial role in maintaining semantics it can also impede the handle points from reaching the target points.\\

\vspace{-11pt}
\noindent\textbf{Enhancing Text Prompt via Prompt Engineering.}
\textit{One straightforward approach to incorporate manipulations into the original text prompt is prompt engineering, allowing direct editing of text prompt to fit the user's intention.}

To validate the effectiveness of prompt engineering~\cite{PE}, we craft an \textit{intention text} prompt that describes the user's intention to edit the image, such as ``\texttt{a photo of a short neck jug and a glass}'', and apply it during the denoising process, as illustrated in \cref{fig:motiv}.
However, this approach encounters the same issue as the original prompt due to the weak coupling between image and text embeddings. 
Thus, we linearly interpolate the intention text embedding with the original text embedding during drag editing.
Nevertheless, actively reflecting the gradual changes in the image embedding remains limited. 
Consequently, we aimed to develop a more sophisticated text embedding modification method in parallel with the gradual image manipulations.
A detailed description of the methods and results of this analysis can be found in \cref{suppsec:pe}.

\noindent\textit{\textbf{Remark.}} Prompt engineering alone may not be sufficient to overcome the limitations of the original text prompt.\\

\noindent\textbf{Necessity for Text Embedding Optimization.}
Based on this series of findings, we advocate for the necessity of alternative text embedding that can: 1) successfully reach the target point while maintaining semantics; 2) by gradually reflect changes introduced during image editing; 3) thus maintain pairing with the edited image embedding to ensure proper denoising.
Furthermore, rather than relying on heuristic and inconsistent methods such as prompt engineering, we propose an optimization approach that is intrinsically parallel to image editing and can be strongly coupled with the edited image.
\section{Method}\label{sec:method}

In this study, we propose \textsc{DragText}, a novel framework to optimize text embedding along with point-based image editing.
We briefly explain the diffusion process first (\cref{subsec:method1}), and then describe the drag process in detail, including text embedding optimization (\cref{subsec:method2}).
Representative diffusion model-based dragging methods~\cite{DragDiffusion, DragNoise, FreeDrag, GoodDrag} share the same approach for motion supervision and point tracking based on DragGAN~\cite{DragGAN} and DragDiffusion~\cite{DragDiffusion}. 
Therefore, we explain \textsc{DragText} primarily with reference to DragDiffusion~\cite{DragDiffusion}, with minor modifications for others detailed in \cref{suppsec:detail}.

\subsection{Preliminaries on Diffusion Models}\label{subsec:method1}
Diffusion models generate images through a forward process and a reverse process. 
By using a denoising U-Net, these models predict noise at a specific time-step $t$ and progressively denoise it to create an image. 
Unlike DDPM~\cite{DDPM}, DDIM~\cite{DDIM} allows for deterministic image generation, making real image editing possible.

Recent developments in Latent Diffusion Models (LDM)~\cite{StableDiffusion} have advanced the diffusion processes to occur efficiently within the latent space.
The input image $\mathbf{x}_0$ is encoded via VAE encoder~\cite{vae} to form a latent vector $\mathbf{z}_0$, which then undergoes a forward process to transition into a noisy latent vector $\mathbf{z}_t$.
This transformation facilitates image editing through the manipulation of this latent vector.
Furthermore, cross-attention mechanisms provide conditioning capabilities with diverse prompts, making text-based conditioning feasible with text embedding $\mathbf{c}$. 
The $\mathbf{z}_t$ is subsequently denoised by using the predicted noise $\mathbf{\epsilon_\theta}(\mathbf{z}_t, t,\mathbf{c})$ from a denoising U-Net, and finally, VAE decoder reconstructs the denoised latent vector into an image $\hat{\mathbf{x}}_0$. 
In our work, we use the pre-trained LDM Stable Diffusion~\cite{StableDiffusion}, which is employed in previous studies~\cite{DragDiffusion, DragNoise, FreeDrag, GoodDrag}.

\subsection{Drag Editing}\label{subsec:method2}
As shown in \cref{fig:pipeline}, we optimize the latent vector $\mathbf{z}_t$ and the text embedding $\mathbf{c}$ according to $n$ number of instruction points $\{h_i = (x_i, y_i), \ g_i = (\tilde{x}_i,\ \tilde{y}_i): i = 1, 2, \ldots, n\}$ via three stages: motion supervision, text optimization, and point tracking.
These stages are sequentially repeated until all handle points have reached corresponding target points or the maximum number of iteration steps $K$ is achieved.

\begin{figure*}[htb!]
    \centering
    \includegraphics[width=\textwidth]{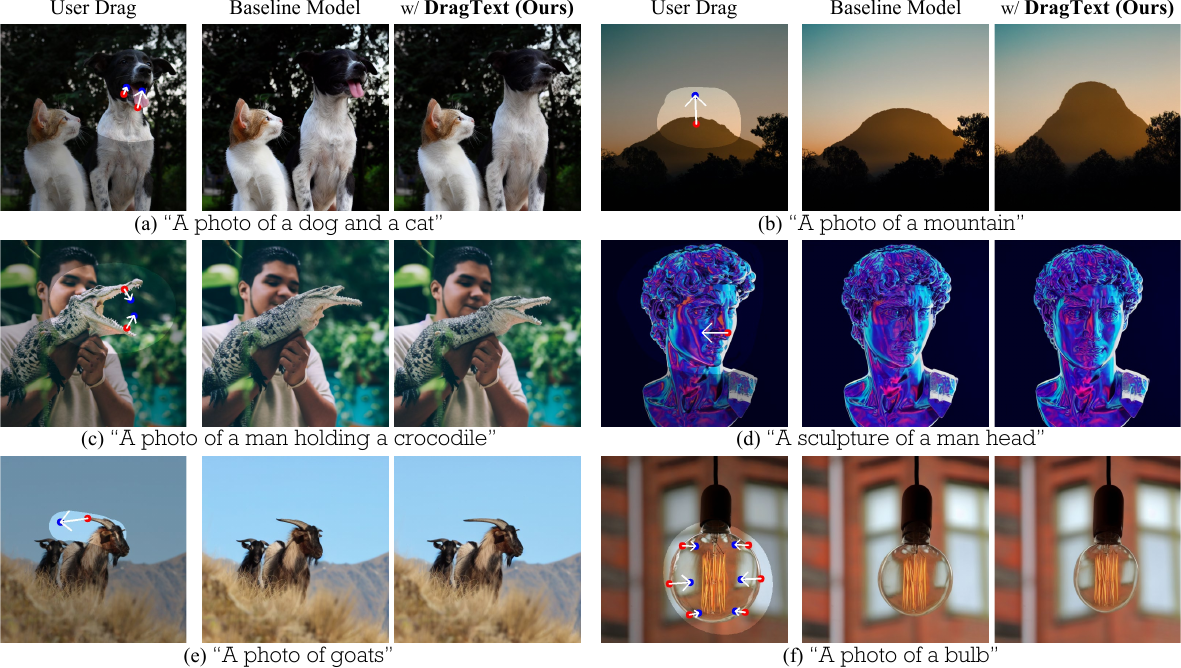}
    \vspace{-18pt}
    \caption{Qualitative results of \textsc{DragText}.
    As demonstrated in the comparison of editing results with and without the application of \textsc{DragText} to the baseline model, DragDiffusion~\cite{DragDiffusion} integrating \textsc{DragText} demonstrates improved semantic control and precision in dragging.}
    \label{fig:main}
    \vspace{-10pt}
\end{figure*}

\subsubsection{Motion Supervision}\label{subsec:method_ms}
\vspace{-5pt}
Motion supervision drags image content around handle points $h_i^k$ toward target points $g_i$ on the image feature space by optimizing the latent vector $\hat{\mathbf{z}}^k_t$, where $k$ implies the number of drag editing iterations.
The feature map $\mathcal{F}(\hat{\mathbf{z}}^k_t, \hat{\mathbf{c}}^k)$ is obtained from the third decoder block of the denoising U-Net, and its corresponding feature vector at the specific pixel location $q$ is denoted as $\mathcal{F}_q(\hat{\mathbf{z}}^k_t, \hat{\mathbf{c}}^k)$.

The motion supervision loss at the $k$-th iteration is defined as:
\begin{equation}
\begin{aligned}
    \mathcal{L}_{\text{ms}} = 
    & \sum_{i=1}^n \sum_{q_1} \left\| \mathcal{F}_{q_1 + d_i}(\hat{\mathbf{z}}_t^k, \hat{\mathbf{c}}^k) - \text{sg}(\mathcal{F}_{q_1}(\hat{\mathbf{z}}_t^k, \hat{\mathbf{c}}^k))\right\|_1 \\
    & + \lambda_\text{image}\left\|\left(\hat{\mathbf{z}}_{t-1}^{k} - \text{sg}(\hat{\mathbf{z}}_{t-1}^0) \right)\odot (\mathbbm{1} - M_\text{image}) \right\|_1,
\end{aligned}
\end{equation}
where $\text{sg}(\cdot)$ is the stop-gradient operator to prevent the gradient from being backpropagated.
$q_1 = \Omega(h_i^k, r_1) = \{(x, y) : |x - x_i^k| \leq r_1, |y - y_i^k| \leq r_1\}$ describes the square region centered at $h_i^k$ with radius $r_1$. 
$d_i = \left(g_i - h_i^k\right)/\left\|g_i - h_i^k\right\|_2$ is the normalized vector from $h^k_i$ to $g_i$ which enables $\mathcal{F}_{q_1 + d_i}$ to be computed by the bilinear interpolation. 
The first term allows $h^k_i$ to be moved to $h^k_i+d_i$, but not the reverse with $\text{sg}(\cdot)$.
$M_\text{image}$ is the binary mask drawn by the user defining an editable region. 
The second term adjusts the extent to which regions outside the mask remain unchanged with $\lambda_\text{image}$ during the optimization. 

For each iteration $k$, $\hat{\mathbf{z}}_t^k$ undergoes a gradient descent step to minimize $\mathcal{L}_{\text{ms}}$:
\begin{equation}
    \hat{\mathbf{z}}_t^{k+1} = \hat{\mathbf{z}}_t^k - \eta_\text{ms}\frac{\partial \mathcal{L}_{\text{ms}}(\hat{\mathbf{z}}_t^k, \hat{\mathbf{c}}^k)}{\partial\hat{\mathbf{z}}_t^k},
\end{equation}
where $\eta_\text{ms}$ is the learning rate for the latent optimization.

\begin{figure*}[t]
    \centering
    \includegraphics[width=\textwidth]{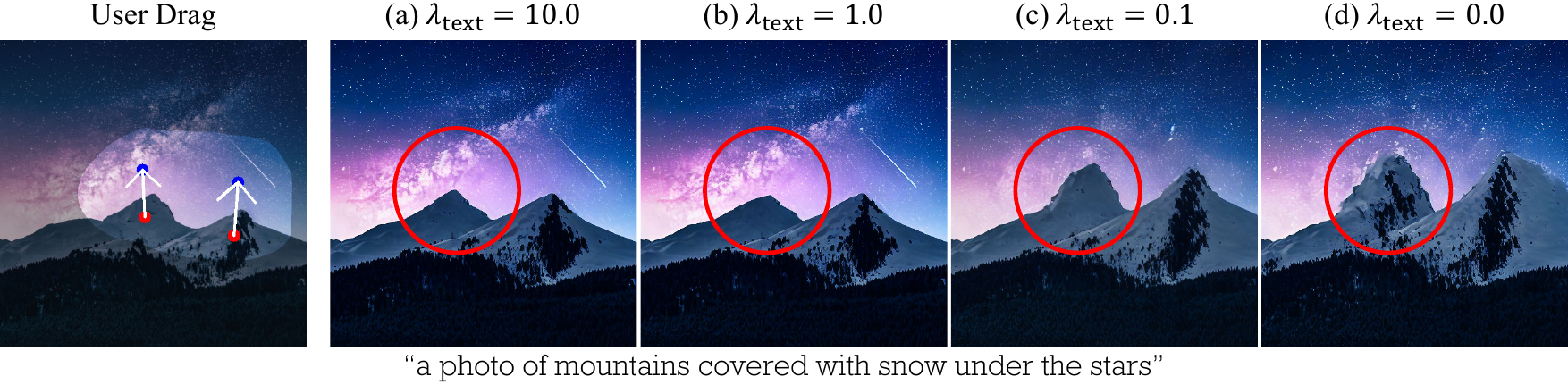}
    \vspace{-18pt}
    \caption{Ablation study on $\lambda_\text{text}$ from text regularization. 
    Larger $\lambda_\text{text}$ values maintain semantics but do not facilitate dragging effectively. 
    As $\lambda_\text{text}$ decreases, dragging becomes more dominant at the expense of maintaining semantics.}    
    \label{fig:regularization}
\vspace{-13pt}
\end{figure*}

\subsubsection{Text Optimization}\label{subsec:method_text}
\vspace{-5pt}
The input text is encoded to the text embedding $\mathbf{c}\in\mathbb{R}^{77\times d}$ through the CLIP text encoder~\cite{CLIP}.
$77$ represents the total number of text tokens, obtained when text is processed by the tokenizer, with only $l$ tokens carrying semantic meaning. 
We use the same loss function as in the motion supervision to optimize the text embedding $\mathbf{\hat{c}}^k$ for ensuring consistency and generalizability.
It allows $\hat{\mathbf{c}}^k$ to follow the latent optimization towards the dragging direction.
In addition, to preserve important content in the original text embedding $\mathbf{c}^0$, we use the mask $M_\text{text}\in\mathbb{R}^{l\times d}$ and regularize the text optimization process.

The text embedding optimization loss at the $k$-th iteration is defined as:
\begin{equation}
\begin{aligned}
    \mathcal{L}_{\text{text}} = 
    & \sum_{i=1}^n \sum_{q_1}\left\| \mathcal{F}_{q_1 + d_i}(\hat{\mathbf{z}}_t^k, \hat{\mathbf{c}}^k) - \text{sg}(\mathcal{F}_{q_1}(\hat{\mathbf{z}}_t^k, \hat{\mathbf{c}}^k)) \right\|_1 \\
    & + \lambda_{\text{text}} \left\| \left(\hat{\mathbf{c}}^k-\text{sg}(\mathbf{c}^0)) \right) \odot M_\text{text} \right\|_1.
\end{aligned}
\end{equation}

For each iteration, $\hat{\mathbf{c}}^k$ undergoes a gradient descent step to minimize $\mathcal{L}_{\text{text}}$:
\begin{equation}
    \hat{\mathbf{c}}^{k+1} = \hat{\mathbf{c}}^k - \eta_\text{text}\cdot\frac{\partial \mathcal{L}_{\text{text}}(\hat{\mathbf{z}}_t^k, \hat{\mathbf{c}}^k )}{\partial \hat{\mathbf{c}}^k },
\end{equation}
where $\eta_\text{text}$ is the learning rate for the text optimization.

\subsubsection{Point Tracking}\label{subsec:method_pt}
After the latent vector $\hat{\mathbf{z}}_t^k$ and text embedding $\hat{\mathbf{c}}^k$ are optimized at $k$-th iteration step, the positions of the handle point $h_i^k$ should change according to the content dragging.
Thus, it is necessary to track the new handle point $h_i^{k+1}$ in the updated feature map $\mathcal{F}(\hat{\mathbf{z}}_t^{k+1}, \hat{\mathbf{c}}^{k+1})$.
To find the new handle points $h_i^{k+1}$ corresponding to the previous handle points $h_i^k$, we find the region in the feature map $\mathcal{F}(\hat{\mathbf{z}}_t^{k+1}, \hat{\mathbf{c}}^{k+1})$ that is most similar to the region around the initial handle points $h_i^0$ in the feature map $\mathcal{F}(\mathbf{z}_t^0, \hat{\mathbf{c}}^0)$:
\begin{equation}
    h_i^{k+1} = \mathop{\arg\min}_{q_2} \left\| F_{q_2}(\hat{\mathbf{z}}_t^{k+1}, \hat{\mathbf{c}}^{k+1}) - F_{h_i^0}(\mathbf{z}^0_t, \mathbf{c}^0) \right\|_1,
\end{equation}
where $q_2$ denotes a point within the region $\Omega(h_i^k, r_2)$. 
It ensures that handle points are consistently updated via nearest neighbor search for subsequent optimization iterations.

\section{Experiments}\label{sec:exp}
To evaluate the effectiveness of \textsc{DragText}, we conducted a series of experiments on DragDiffusion~\cite{DragDiffusion}, FreeDrag~\cite{FreeDrag}, DragNoise~\cite{DragNoise}, and GoodDrag~\cite{GoodDrag}.
The proposed method is applicable to these without hyperparameter tuning, ensuring robustness.
Moreover, text optimization hyperparameters are uniformly applied across all methods.

Specifically, we set $\lambda_\text{text}=0.1$ to match the image regularization factor $\lambda_\text{image}=0.1$, and the text optimization learning rate $\eta_\text{text}=0.004$. 
Detailed information about implementation is in \cref{suppsec:implement}.
Our method requires no additional GPU resources beyond those used by baseline methods and is executed using a single NVIDIA RTX A6000.
Qualitative results are based on DragDiffusion, and further examples using other methods are in \cref{suppsec:result}.

\begin{table}[t]
    \caption{Quantitative results on the DragBench dataset~\cite{DragDiffusion}}
    \vspace{-13pt}
    \centering
    \vspace{2mm} 
    \scalebox{0.7}{
    \renewcommand{\arraystretch}{1.1} 
    \setlength{\tabcolsep}{4pt} 
    \begin{tabular}{lcccccccc}
    \toprule
    Methods & \multicolumn{2}{c}{DragDiffusion} & \multicolumn{2}{c}{FreeDrag} & \multicolumn{2}{c}{DragNoise} & \multicolumn{2}{c}{GoodDrag} \\ \midrule
    Metric & LPIPS & MD & LPIPS & MD & LPIPS & MD & LPIPS & MD\\ \midrule
    Baseline & \textbf{0.117} & 33.21 & \textbf{0.101} & 34.44 & 0.103 & 35.55 & \textbf{0.129} & 24.29 \\
    \rowcolor[HTML]{eff5fe} 
    + \textsc{DragText} & 0.124 & \textbf{29.78} & 0.104 & \textbf{31.59} & \textbf{0.097} & \textbf{34.23} & 0.131 & \textbf{20.53} \\
    \bottomrule
    \end{tabular}
    }
    \label{tab:quantitative}
    \vspace{-3mm} 
\end{table}

\begin{figure}[t]
    \centering
    \includegraphics[width=0.47\textwidth]{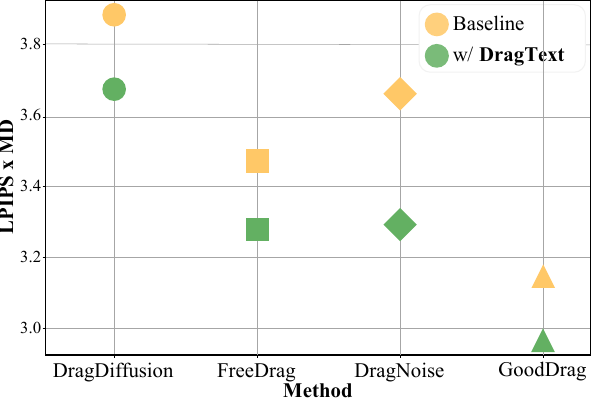}
    \vspace{-10pt}
    \caption{Quantitative results comparing the product of LPIPS and MD for baseline methods and w/ \textsc{DragText}. 
    Performance improved across all methods when applying \textsc{DragText}.}
    \vspace{-15pt}
    \label{fig:LPIPSxMD}
\end{figure}

\subsection{Qualitative Evaluation}
\vspace{-5pt}
\begin{figure*}[t]
    \centering
    \includegraphics[width=\textwidth]{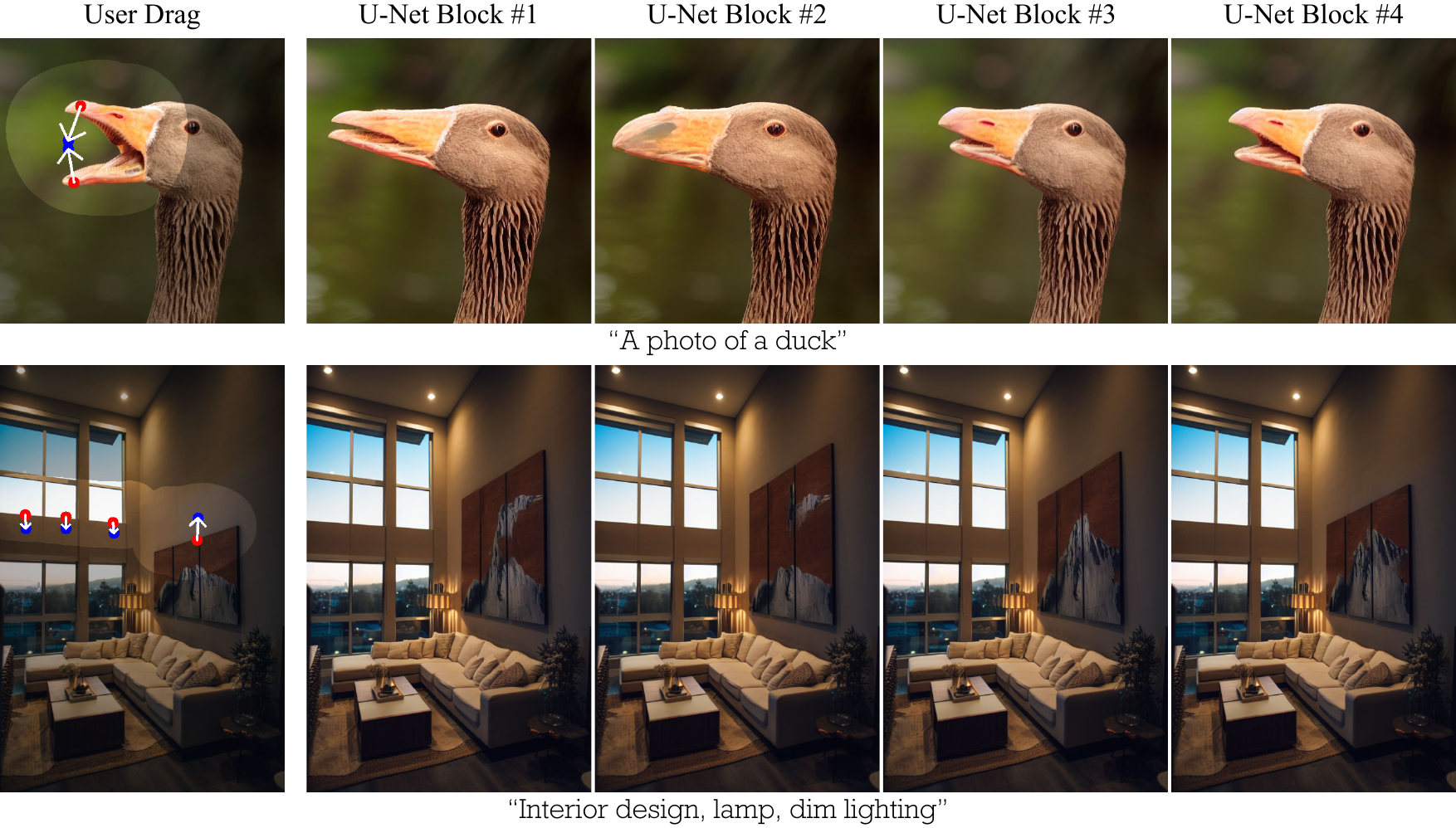}
    \vspace{-20pt}
    \caption{Ablation study on the block number of U-Net decoder, which determines the location where the feature map is obtained. 
    Lower U-Net blocks (\eg, Block 1) focus on low semantic information, while higher U-Net blocks (\eg, Block 4) concentrate on high semantic information. 
    The best results are observed in block 3, where both types of information are appropriately mixed.}
    \vspace{-10pt}
    \label{fig:unet}
\end{figure*}

In \cref{fig:main}, we compared qualitative results of methods with and without \textsc{DragText} on the DragBench dataset~\cite{DragDiffusion}.
Existing methods struggled to reach target points or suffer from the degradation of important image semantics, due to the use of image and text embeddings that are not strongly coupled.
In contrast, by applying \textsc{DragText}, images can be accurately dragged to the specified target points (\eg, (b) mountain peak and (d) sculpture) while preserving essential semantics that should not disappear (\eg, (c) the mouth of a crocodile). 
Furthermore, \textsc{DragText} effectively manages content removal and creation caused by dragging, producing high-quality results  (\eg (a) the tongue of a dog and (e) the horns of a goat). 

\subsection{Quantitative Evaluation}
\vspace{-5pt}
In \Cref{tab:quantitative}, we present the quantitative comparison of methods with and without \textsc{DragText} on the DragBench dataset~\cite{DragDiffusion}.
The evaluation metrics are LPIPS~\cite{LPIPS} and mean distance (MD)~\cite{DragGAN}.
LPIPS measures the perceptual similarity between the edited image to the original image.
MD measures the distance between the final position of the handle point $h$ and the target point $g$ in the feature map of the DIFT~\cite{DIFT}.
While baseline methods and \textsc{DragText} applied methods show little differences in LPIPS, there are significant improvements in MD. 
As shown in \cref{fig:LPIPSxMD}, differences are more dramatic, when considering LPIPS and MD simultaneously. 
Applying \textsc{DragText} results in substantial performance improvements across all methods, indicating that text embedding optimization allows for effective dragging while preserving image fidelity.

\subsection{Ablation Study}\vspace{-5pt}
\noindent\textbf{Effect of Text Regularization.}
In \cref{fig:regularization}, we investigated the impact of $\lambda_\text{text}$ designed to preserve the semantics of the image and text. 
We edited images by varying $\lambda_\text{text}$ from $0$ to $10$, which controls the degree of regularization for the text embedding. 
As $\lambda_\text{text}$ approaches $0$, the force of dragging becomes more dominant than the preservation of semantics. 
For example, in \cref{fig:regularization} (d), the icecap has disappeared, indicating extensive dragging. 
Conversely, applying an excessively high $\lambda_\text{text}$ during text embedding optimization maintains the original content of the image and inhibits editing. 
In \cref{fig:regularization} (a), minimal dragging is observed, showing that the original image remains largely unchanged. 
Our experiments showed that setting $\lambda_\text{text}$ to $0.1$ achieves the optimal balance between dragging and content preservation.
This value can be adjusted based on users' editing preferences.

\begin{figure*}[t]
    \centering
    \includegraphics[width=\textwidth]{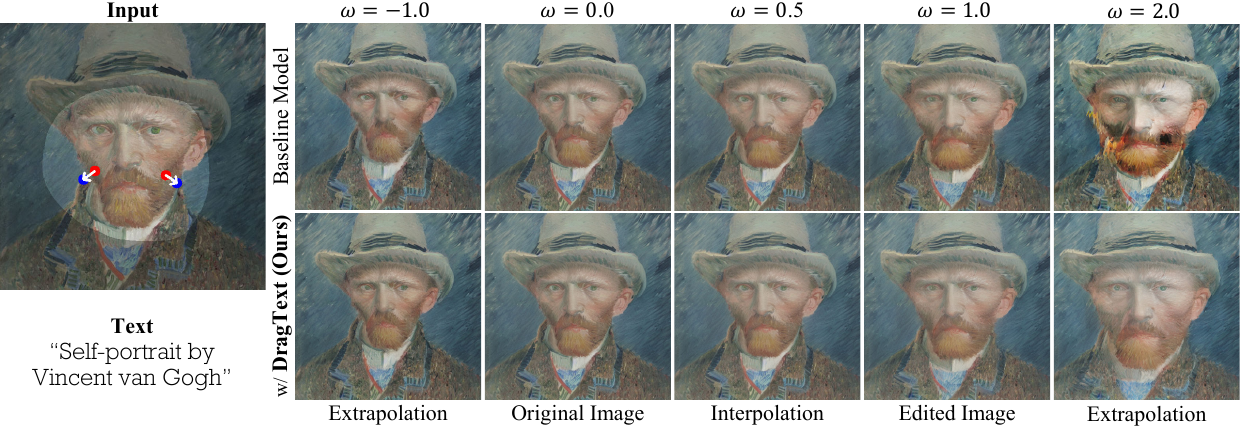}
    \vspace{-23pt}
    \caption{Changes in images when manipulating the image and text embeddings after optimization finished. 
    $\omega$ is a factor of the linear interpolation between the original embeddings and the optimized embeddings. 
    As $\omega$ approaches $0$, the image resembles the original image; as $\omega$ approaches $1$, the image resembles the dragged image. 
    \textsc{DragText} also generates images outside the original editing range (e.g., extrapolating to $\omega = -1.0$ and $\omega = 2$) in the drag direction, preserving the semantic content at the same time.}
    \label{fig:interpolation}
    \vspace{-13pt}
\end{figure*}

\vspace{3pt}
\noindent\textbf{Effect of the Block Number of U-Net Decoder.}
In \cref{fig:unet}, we perform \textsc{DragText} using feature maps from four different U-Net decoder blocks to assess the impact of each block on text embedding.
The image embedding used the feature map from the 3rd block, as in DragDiffusion ~\cite{DragDiffusion}. 
Images optimized using the feature map from the 3rd block exhibited optimal semantic preservation and effective dragging. 
Feature maps from lower blocks (\eg Block 1) are hard to maintain semantics, whereas feature maps from higher blocks (\eg Block 4) preserve semantics well but result in poor dragging performance. 
This phenomenon is likely due to the lower blocks of the U-Net containing low-frequency information of the image~\cite{FreeU}. 
Quantitative results in \Cref{suppsec:unet} support our qualitative evaluation.
\section{Manipulating Embeddings}\label{sec:app}
\vspace{-5pt}

\textsc{DragText} enables the model to comprehend the degree and direction of changes in the image while preserving the important semantic content.
In \cref{fig:interpolation}, we applied the same linear interpolation to both the image and text embeddings before and after editing to generate intermediate images. 
The image and text embeddings of intermediate images are defined as follows:
$\hat{\mathbf{z}} = (1 - \omega) \mathbf{z}_0 + \omega \hat{\mathbf{z}}_t^k, \quad
\hat{\mathbf{c}}= (1 - \omega) \mathbf{c}^0 + \omega \hat{\mathbf{c}}^k.$
When $\omega=0$, the result corresponds to the original image, and as $\omega$ approaches $1$, it gets closer to the dragged image. 
Remarkably, this phenomenon is maintained even for $\omega \notin [0, 1]$, allowing the degree of dragging to be adjusted even beyond the editing process.
With \textsc{DragText}, simple manipulations preserved semantic content while allowing control over the degree and direction of the drag, as \textsc{DragText} optimizes the text and image embedding together. 
In contrast, the baseline model failed to maintain semantics as the text embedding necessary for semantic preservation is not jointly optimized.

\section{Limitation}\label{sec:limits}
\begin{figure}[t]
\vspace{-4pt}
    \centering
    \includegraphics[width=0.47\textwidth]{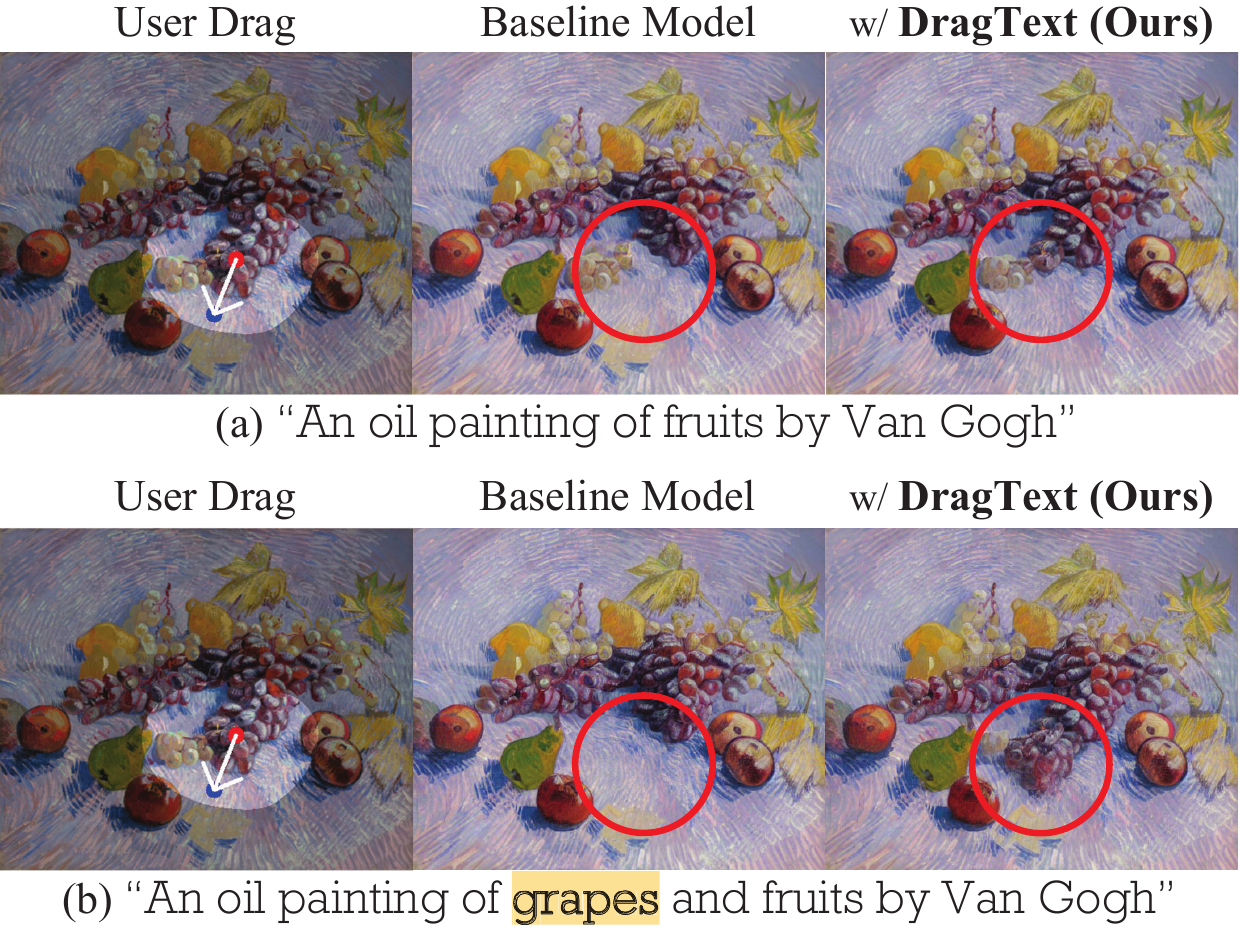}
    \vspace{-5pt}
    \caption{
    Limitations of diffusion-based editing methods. 
    (a) The diffusion-based methods sometimes lose important features during the drag optimization process, resulting in content loss, especially when the context prompt lacks specificity. 
    (b) However, if the user provides precise content information, specifying ``grapes'' as the target to be dragged, \textsc{DragText} successfully performed the drag while preserving the content, unlike the baseline model.}
    \vspace{-16pt}
    \label{fig:limitation}
\end{figure}

\vspace{-5pt}
Diffusion-based editing methods have the common problem of often losing the dragged feature in the latent space, leading to the vanishing of content in surrounding objects that are not the primary editing target.
Specifically, in \cref{fig:limitation} (a), grape grains within the red circle disappeared when images are dragged, regardless of whether \textsc{DragText} is applied.
Nevertheless, this situation changes dramatically when the user provides a more detailed text prompt.
For instance, in \cref{fig:limitation} (b), when the user provides the exact word ``\texttt{grapes}'' related to the editing target, \textsc{DragText} dragged the grape feature since it optimizes the text embedding along with the image embedding, preserving the original content.
The optimized text embedding plays a complementary role for the features lost in the image embedding space.
On the other hand, the baseline model still could not drag the grape feature.
\section{Conclusion}\label{sec:conclusion}
\vspace{-5pt}

In this study, we introduced \textsc{DragText}, a method emphasizing the critical role of text prompts in point-based image editing with diffusion models. 
Our findings revealed that static text embeddings hinder the editing process by causing drag halting and a loss of semantic integrity. 
By optimizing text embeddings in parallel with image embeddings, \textsc{DragText} ensures that text and image embeddings remain strongly coupled, ensuring better drag accuracy and content preservation. 
Moreover, it integrates seamlessly with existing diffusion-based drag methods, leveraging their strengths while consistently enhancing performance and fidelity in image editing.
Our work underlines the importance of considering text-image interaction in point-based editing, providing a novel insight for future developments.

\let\thefootnote\relax\footnote{\scriptsize\textbf{Acknowledgments.} This work was supported by IITP RS-2024-00457882 (National AI Research Lab Project), NRF RS-2024-00345806, and NRF RS-2023-00219019 funded by Korean Government (MSIT).}

{\small
\bibliographystyle{ieee_fullname}
\bibliography{Drag}
}

\clearpage
\onecolumn
\appendix
\maketitle
\thispagestyle{empty}

\section{Additional Material: Code \& Project Page}\label{suppsec:code}
The code has been submitted as a zip file along with the Supplement.
Our results are presented in an easily accessible format on our project page. 
The link to the project page is as follows:~\url{https://micv-yonsei.github.io/dragtext2025/}

\vspace{1cm}
\section{More Details on Prompt Engineering}\label{suppsec:pe}
In this section, we provide a detailed explanation of the analysis from Section 3.2., examining the effectiveness of prompt engineering in point-based image editing.
First, we explain how the analysis was conducted. 
Next, we present more examples of the intention text we used and the corresponding results.

\vspace{0.5cm}
\subsection{Implementation Details}
\noindent\textbf{Drag Editing with Intention Text.}
We estimated the editing intentions for each image using the handle points, the target points, and the image masks provided by the DragBench dataset~\cite{DragDiffusion}. 
Additionally, we referenced the edited results from four methods~\cite{DragDiffusion, FreeDrag, DragNoise, GoodDrag}.
The intention text prompts were crafted by injecting these editing intentions into the original text prompts. 
To ensure that secondary changes in the text prompts did not affect the editing results, we minimized alterations to the vocabulary and sentence structures of the original text prompts.

For example, consider an image with the original prompt \texttt{"a photo of a jug and a glass"} where the jug's neck needs to be shortened. 
We can craft an intention text prompt via ~\cite{GPT4} such as:

\begin{center}
\noindent\fbox{\begin{varwidth}{\linewidth}\ttfamily
"Create an image of a jug with a shorter neck. Shorten the neck by the distance between a red dot and a blue dot. The jug should have a smooth, glossy finish. Place the jug against a simple, neutral background."
\end{varwidth}}
\end{center}

However, this significantly altered the content of the original text prompt.
This alteration makes it challenging to discern whether the changes in the edited result were due to these secondary modifications or the incorporation of the editing intention in the text. 
Consequently, we incorporated concise terms representing the editing intention while preserving the original vocabulary and sentence structure as much as possible, for example: \texttt{"a photo of a \emph{short-neck} jug and a glass."}
\\

\noindent\textbf{Linear Interpolation.}
To reflect gradual changes in the image embeddings to the text embeddings, we linearly interpolate between the original text embeddings and the intention text embeddings during the dragging process. 
The weights of the original text embeddings and the intention text embeddings are determined based on the distance between the handle point $h^k_i$ and the target point $g_i$:

\begin{equation*}
    w^k = \frac{\sum_{i=1}^n \left\|g_i - h^k_i\right\|_2}{\sum_{i=1}^n \left\|g_i - h^0_i\right\|_2}
\end{equation*}
\vspace{0pt}
\begin{equation*}
    \hat{\mathbf{c}}^k 
    = w^k \cdot \hat{\mathbf{c}}^k_\text{org} + (1-w^k) \cdot \hat{\mathbf{c}}^k_\text{int}
\end{equation*}
where $\hat{\mathbf{c}}^k_\text{org}$ is the original text embedding and $\hat{\mathbf{c}}^k_\text{int}$ is the intention text embedding.
At the beginning of the point-based editing, $w^k=1$ so $\hat{\mathbf{c}}^k = \hat{\mathbf{c}}^k_\text{org}$.
Conversely, as the dragging progresses and handle points approach target points, the weight value $w^k$ increases, resulting in a higher proportion of $\hat{\mathbf{c}}^k_\text{int}$.
In this way, as the image is progressively edited, the proportion of the intention text embedding is gradually increased. 

\clearpage
\subsection{More Qualitative Results for Prompt Engineering}
\begin{figure*}[hb!]
    \centering
    \includegraphics[width=\textwidth]{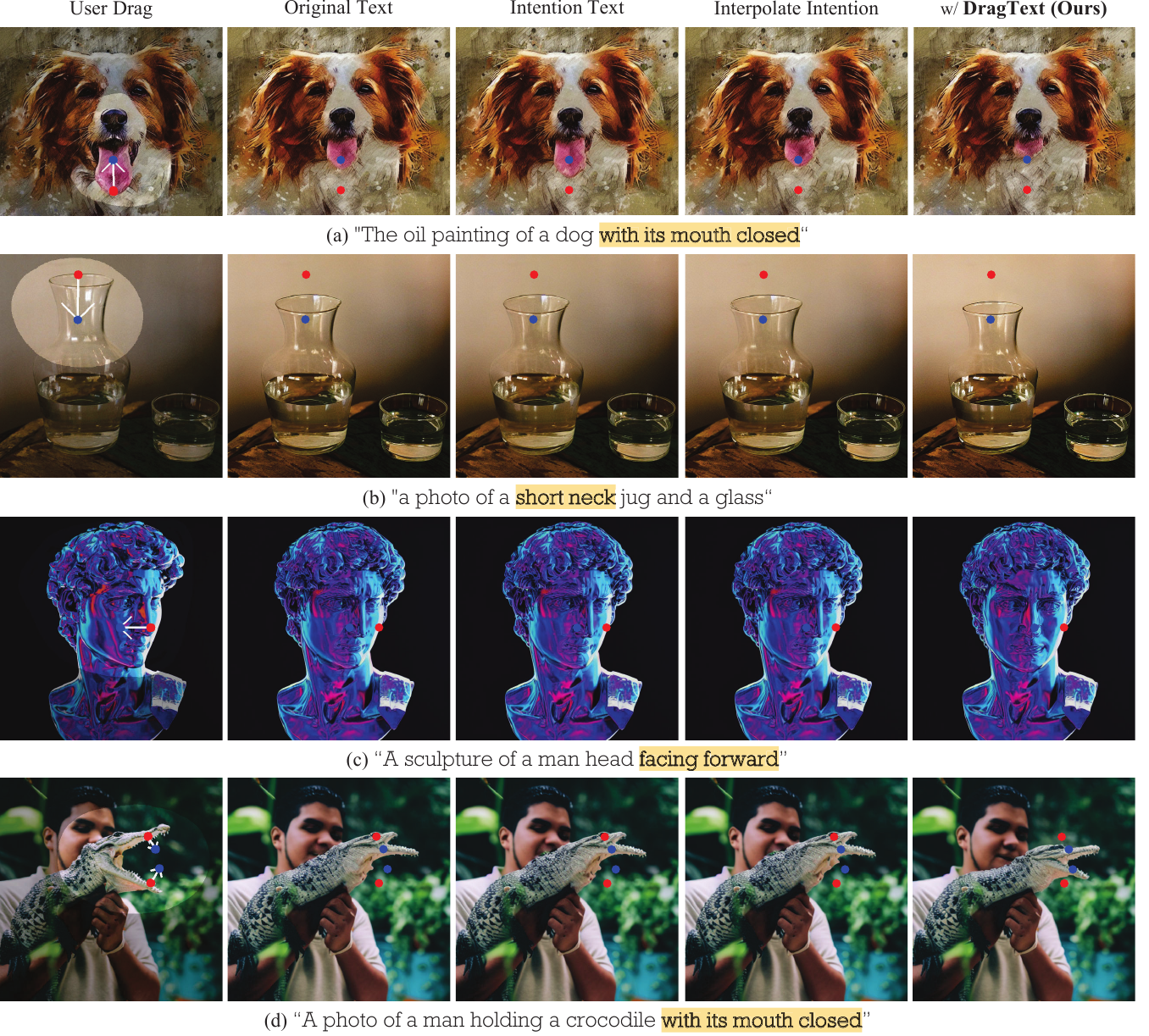}
    \vspace{-15pt}
    \caption{\textbf{More qualitative results for prompt engineering (\ie the intention text, and interpolated intention text).}
    Red points and blue points represent handle points and their target points respectively.
    Overall, \textsc{DragText} moved the content at the handle point of the User Drag image closer to the target point. 
    Additionally, in (b), it was observed that interpolating the intention text with the original text makes it challenging to maintain the object's style.
    In (d), \textsc{DragText} preserved semantics better than other methods while dragging appropriately.
    }
    \label{suppfig:pe}
    \vspace{-10pt}
\end{figure*}

In \cref{suppfig:pe}, we present additional results for prompt engineering.
Corresponding results for \textsc{DragText} are also presented to validate the effectiveness of our approach in comparison to prompt engineering.
Prompt engineering was found to have little impact on alleviating drag halting. 
In contrast, \textsc{DragText} dragged handle points closer to target points, compared to the original text prompt, the intention text prompt, and their interpolation.

\clearpage
\section{More Details on \textsc{DragText}}\label{suppsec:detail}
In this section, we provide a comprehensive overview of our method to ensure clarity and ease of understanding. 
We describe the pipeline of \textsc{DragText} using pseudo-code to aid in understanding our approach. 
Additionally, we detail the modifications necessary to apply \textsc{DragText} to other point-based image editing methods. 

\vspace{0.5cm}
\subsection{Pseudo Code of \textsc{DragText}}
\begin{algorithm}[h!]
    \caption{Pipeline of \textsc{DragText}}
    \algorithmicrequire{
        Input image $\mathbf{x}_0$, text prompt $\mathbf{c}$, image mask $M_\text{image}$,  handle points $\{{h}_{i}\}_{i=1}^n$, target points $\{{g}_{i}\}_{i=1}^n$, 
        denoising U-Net $U_\theta$, diffusion time step $t$, maximum number of iterations steps $K$ \\
    }
    \algorithmicensure{
    Output image $\hat{\mathbf{x}}_0$}
    \begin{algorithmic}[1]
    \State $\mathbf{z}_t \gets$ apply DDIM inversion to $\mathbf{x}_0$ conditioned on $\mathbf{c}$
    \State $\mathbf{z}_t^0, \mathbf{c}^0, h_i^0 \gets \mathbf{z}_t, \mathbf{c}, h_i$
    \For{$k$ in $0:K-1$}
        \State $\mathcal{F}(\hat{\mathbf{z}}_t^k, \hat{\mathbf{c}}^k) \gets U_{\theta}(\hat{\mathbf{z}}_t^k;\hat{\mathbf{c}}^k)$
        \State Update $\hat{\mathbf{z}}_t^k$ using motion supervision
        \State Update $\hat{\mathbf{c}}^k$ using text optimization
        \State $\hat{\mathbf{z}}_t^{k+1}, \hat{\mathbf{c}}^{k+1} \gets \hat{\mathbf{z}}_t^k, \hat{\mathbf{c}}^k$
        \State Update $\{h_{i}^{k+1}\}_{i=1}^n$ using points tracking
    \EndFor
    \State $\hat{\mathbf{z}}_t, \hat{\mathbf{c}} \gets \hat{\mathbf{z}}_t^K, \hat{\mathbf{c}}^K$
    \For{$t$ in $t:1$}
        \State $\hat{\mathbf{z}}_{t-1} \gets$ apply denoising to $\hat{\mathbf{z}}_{t}$ conditioned on $\hat{\mathbf{c}}$
    \EndFor
    \State $\hat{\mathbf{x}}_0 \gets \hat{\mathbf{z}}_0$
    \end{algorithmic}
    \label{supp:alg}
\end{algorithm}

In DragText, another important element is the mask $M_\text{text}\in\mathbb{R}^{l\times d}$. 
This mask is used to regularize text embedding optimization but is not an input.
Instead, it is automatically calculated by the CLIP tokenizer~\cite{CLIP}.
After the text prompt passes through the tokenizer, the tokens excluding \texttt{[EOS]} and \texttt{[PAD]} tokens contain significant semantic information.
The length of these important tokens is $l$.

\vspace{0.5cm}
\subsection{Modifications for Integrate with Other Methods}
In the main paper, we explained \textsc{DragText} based on DragDiffusion~\cite{DragDiffusion}. 
We chose this method because representative diffusion model-based dragging methods~\cite{FreeDrag, DragNoise, GoodDrag} all utilize approaches from DragGAN~\cite{DragGAN} and DragDiffusion.
Therefore, they are constructed upon the foundation of DragDiffusion.
However, they also developed techniques to overcome the limitations of DragDiffusion. 
Taking these improvements into account, we made minor modifications to \textsc{DragText} to adapt our approach for each method. 

\vspace{0.25cm}
\subsubsection{FreeDrag}
Point-based image editing has faced challenges, such as the disappearance of handle points during the dragging process. 
Additionally, point tracking often fails to reach the target point because it moves to the location with the smallest difference in the feature map. 
To address these issues, FreeDrag introduces \textit{Template Feature} and restricts the path of point tracking to a straight line.

FreeDrag generates the corresponding template features $\mathcal{T}_{i}^{k}$ for each of the $n$ handle points. 
During the optimization step, the feature map is optimized to match the template features:

\begin{equation*}
\begin{aligned}
\mathcal{L}_{\text{ms}} = 
    & \sum_{i=1}^n \sum_{q_1} \left\| \mathcal{F}_{q_1 + d_i}(\hat{\mathbf{z}}_t^k, \hat{\mathbf{c}}^k) - \mathcal{T}_i^k\right\|_1 \\
    & + \lambda_\text{image}\left\|\left(\hat{\mathbf{z}}_{t-1}^k-\text{sg}(\hat{\mathbf{z}}_{t-1}^0) \right)\odot (\mathbbm{1} - M_\text{image}) \right\|_1.
\end{aligned}
\end{equation*}
This involves up to five iterations of motion supervision until the predefined conditions are met. 
Depending on the outcome, feature adaptation is categorized into (a) well-learned features, (b) features in the process of learning, and (c) poorly learned features. 
Point tracking is then performed based on these categories. 
This process is repeated until the handle points reach the target points, after which the image is denoised to produce the edited image. 

In FreeDrag, if the template feature is poorly learned (category (c)), the point not only reverts to its previous position but also reuses the template feature map without updating its values. 
Inspired by this approach, our DragText computes the text loss only during the (a) and (b) processes to align with the image. In cases categorized as (c), the text embedding is excluded from the optimization process.
Therefore, we define $\alpha_{i}$ as 1 for cases (1) and (2), and 0 for case (3). 
The text loss is then defined as follows:
\begin{equation*}
\begin{aligned}
\mathcal{L}_{\text{text}} = 
    & \sum_{i=1}^n \sum_{q_1} \alpha_{i}\left\| \mathcal{F}_{q_1 + d_i}(\hat{\mathbf{z}}_t^k, \hat{\mathbf{c}}^k) - \mathcal{T}_i^k\right\|_1 \\
    & + \lambda_{\text{text}} \left\| \left(\hat{\mathbf{c}}^k-\text{sg}(\mathbf{c}^0)) \right) \odot M_\text{text} \right\|_1.
\end{aligned}
\end{equation*}
In DragText, during image optimization, $\hat{\mathbf{c}}^k$ does not undergo gradient descent, and during text optimization, the latent vector $\hat{\mathbf{z}}_t^k$does not undergo gradient descent.

\vspace{0.25cm}
\subsubsection{DragNoise}
The bottleneck features $\mathbf{s}_t$ effectively capture richer noise semantics and efficiently capture most semantics at an early timestep $t$.
Thus, DragNoise optimizes the bottleneck feature $\mathbf{s}_t$ of the U-Net instead of the latent vector $\mathbf{z}_t$ thereby shortening the back-propagation chain.
Accordingly, \textsc{DragText} optimizes $\mathbf{s}_t$ instead of $\mathbf{z}_t$ during the image optimization processes.
For each iteration $k$, $\mathbf{\hat{s}}_t^k$ undergoes a gradient descent step to minimize $\mathcal{L}_{\text{ms}}$:
\begin{equation*}
    \mathbf{\hat{s}}_t^{k+1} = \mathbf{\hat{s}}_t^k - \eta_\text{ms}\frac{\partial \mathcal{L}_{\text{ms}}(\mathbf{\hat{s}}_t^k, \hat{\mathbf{c}}^k)}{\partial\mathbf{\hat{s}}_t^k}.
\end{equation*}

In \textsc{DragText}, neither the latent vector $\mathbf{z}_t^k$ nor the bottleneck feature $\mathbf{s}_t^k$ undergoes gradient descent during the text optimization. 
So the text optimization procedure is not modified in DragNoise.

\vspace{0.25cm}
\subsubsection{GoodDrag}
GoodDrag alternates between dragging and denoising, introducing periodic corrections to mitigate accumulated errors.
This approach is different from traditional diffusion-based dragging methods. 
They generally execute all drag operations at once before denoising the optimized noisy latent vector $\hat{\mathbf{z}}_t$. 
During the denoising process, which involves sampling images $\hat{\mathbf{x}}_0$ from a noisy latent vector $\hat{\mathbf{z}}_t$, perturbations from dragging are corrected.
However, if the denoising process is performed only after all drag operations are completed, the errors accumulate too significantly to be corrected with high fidelity.
To address this, GoodDrag applies one denoising operation after $B$ image optimization and point tracking steps.

For example, the latent vector $\hat{\mathbf{z}}^k_t$ has been denoised $\lfloor\frac{k}{B}\rfloor$ times, the drag optimization is performed at the timestep $t = T - \frac{k}{B}$. 
To ensure this process is consistent, the total number of drag steps $K$ should be divisible by $B$.
Since \textsc{DragText} performs one text optimization step after one image optimization step, we sequentially repeat the image optimization, text optimization, and point tracking steps $B$ times, and then apply one denoising operation.

Moreover, when drag editing moves the handle points $\{{h}_{i}\}_{i=1}^n$, the features around handle points tend to deviate from their original appearance. 
This deviation can lead to artifacts in the edited images and difficulties in accurately moving the handle points.
To prevent this, GoodDrag keeps the handle point $h^k_i$ consistent with the original point $h^0_i$, throughout the entire editing process:

\begin{equation*}
\begin{aligned}
    \mathcal{L}_{\text{ms}} = 
    & \sum_{i=1}^n \sum_{q_1} \left\| \mathcal{F}_{q_1 + 4\cdot d_i}(\hat{\mathbf{z}}_t^k, \hat{\mathbf{c}}^k) - \text{sg}(\mathcal{F}_{q_1^0}(\hat{\mathbf{z}}_t^0, \hat{\mathbf{c}}^0))\right\|_1 \\
    & + \lambda_\text{image}\left\|\left(\hat{\mathbf{z}}_{t-1}^{k} - \text{sg}(\hat{\mathbf{z}}_{t-1}^0) \right)\odot (\mathbbm{1} - M_\text{image}) \right\|_1,
\end{aligned}
\end{equation*}
where $q_1 = \Omega(h_i^k, r_1)$, and $q_1^0$ describes the square region centered at the original handle point $h_i^0$. And, drag operations per denoising step $B=10$.
Similarly, \textsc{DragText} ensures the handle point $h^k_i$ remains consistent with the original point $h^0_i$ during text optimization:
\begin{equation*}
\begin{aligned}
    \mathcal{L}_{\text{text}} = 
    & \sum_{i=1}^n \sum_{q_1}\left\| \mathcal{F}_{q_1 + 4\cdot d_i}(\hat{\mathbf{z}}_t^k, \hat{\mathbf{c}}^k) - \text{sg}(\mathcal{F}_{q_1^0}(\hat{\mathbf{z}}_t^0, \hat{\mathbf{c}}^0)) \right\|_1 \\
    & + \lambda_{\text{text}} \left\| \left(\hat{\mathbf{c}}^k-\text{sg}(\mathbf{c}^0)) \right) \odot M_\text{text} \right\|_1.
\end{aligned}
\end{equation*}

Additionally, GoodDrag faced increased optimization difficulty from this design, due to the larger feature distance compared to the original motion supervision loss. 
To mitigate this, a smaller step size and more motion supervision steps are used for optimization. 
This strategy is also applied in \textsc{DragText}.

\vspace{1cm}
\section{Implementation Details}\label{suppsec:implement}
\begin{table}[h]
    \centering
    \caption{Hyperparameters for point-based image editing methods and \textsc{DragText}.}
    \renewcommand{\arraystretch}{0.8} 
    \resizebox{0.75\columnwidth}{!}{%
    \begin{tabular}{@{}c|cccc@{}}
    \toprule 
    Methods & DragDiffusion  & FreeDrag  & DragNoise & GoodDrag \\ \midrule\midrule
    Diffusion Model & \multicolumn{4}{c}{Stable Diffusion 1.5} \\
    Time Step ($T$) & 35 & 35 & 35 & 38 \\
    LoRA Training Step & 80 & 200 & 200 & 70\\ 
    Maximum Optimization Step ($K$) & 80 & 300 & 80 & 70 \\
    Square radius $r_1$ & 1 & 3 & 1 & 4 \\ \midrule
    
    \multicolumn{5}{c}{Motion Supervision Loss} \\ \midrule
    Learning Rate ($\eta_\text{ms}$) & 0.01 & 0.01 & 0.02 & 0.02 \\
    Optimizer & \multicolumn{4}{c}{Adam} \\ 
    $\lambda_\text{image}$ & 0.1 & 10 & 0.2 & 0.2 \\ \midrule
    
    \multicolumn{5}{c}{+ Text Optimization Loss (\textsc{DragText})} \\ \midrule
    Learning Rate $\eta_\text{text}$ & \multicolumn{4}{c}{0.004} \\
    Optimizer & \multicolumn{4}{c}{Adam} \\
    $\lambda_\text{text}$ & \multicolumn{4}{c}{0.1} \\ \midrule
    
    \multicolumn{5}{c}{Point Tracking} \\ \midrule
    Square radius $r_2$ & 3 & - & 3 & 12 \\
    Drag Optimization per Point Tracking & 1 & - & 1 & 3 \\ \bottomrule
    \end{tabular}}
    \label{supptab:implement}
\end{table}

In \Cref{supptab:implement}, we listed the hyperparameters used for each point-based image editing method~\cite{DragDiffusion, FreeDrag, DragNoise, GoodDrag}. 
These values were consistently used in both the \textit{Baseline} and \textit{w/ \textsc{DragText}} experiments. 
For a fair comparison, we applied the same hyperparameter values from the respective paper to our experiments. 
Additionally, we maintained the same text optimization loss across all methods to demonstrate the robustness of our approach.

In FreeDrag, values related to point tracking are omitted since it replaces point tracking with line search.

\vspace{1cm}
\section{More Qualitative Results}\label{suppsec:result}
\begin{figure*}[htb!]
    \centering
    \includegraphics[width=\textwidth]{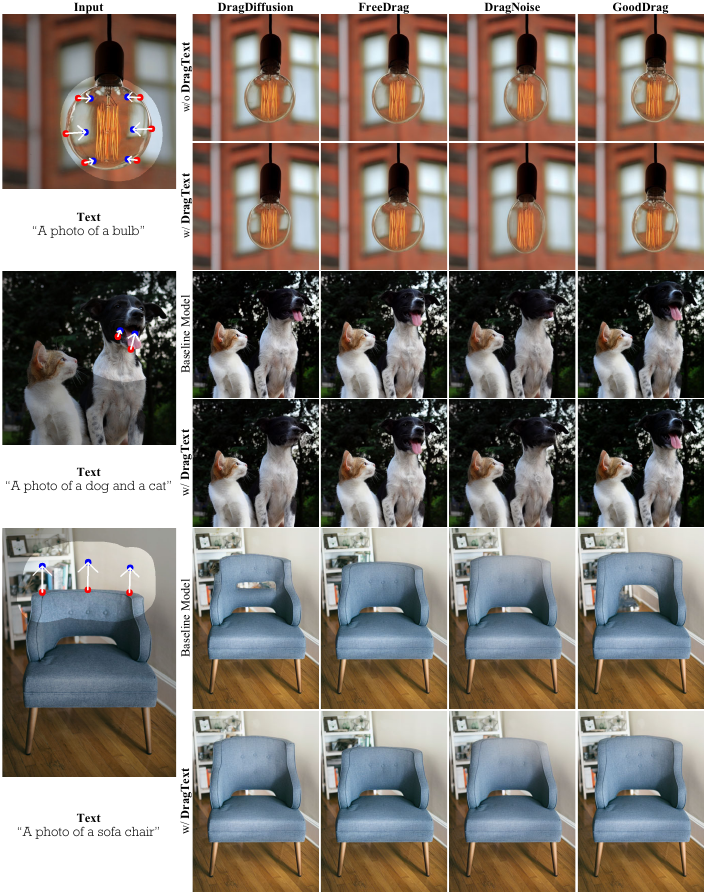}
    \vspace{-15pt}
    \caption{
    \textbf{\textsc{DragText} effectively moved the handle point to the target point while preserving semantics.}
    Moreover, the consistent performance across different methods demonstrates the generalizability of \textsc{DragText}.
    }
    \label{suppfig:main}
    \vspace{-10pt}
\end{figure*}

In \cref{suppfig:main}, we additionally present the results of applying \textsc{DragText} to each method~\cite{DragDiffusion, FreeDrag, DragNoise, GoodDrag}.
In our experiments, we applied \textsc{DragText} to various point-based image editing methods and evaluated their performance.
The results show that \textsc{DragText} can effectively drag the handle points to their corresponding target points while maintaining the semantic integrity of the original image.
Moreover, the consistent success of \textsc{DragText} across multiple methods underscores its robustness and adaptability. 

\clearpage

\section{Evaluation Metrics}\label{suppsec:metric}
\vspace{0.5cm}
\subsection{LPIPS}
LPIPS~\cite{LPIPS} uses ImageNet classification models such as VGG~\cite{VGG}, SqueezeNet~\cite{squeezenet}, and AlexNet~\cite{alexnet}. 
We measured LPIPS using AlexNet. 
LPIPS measures the similarity between two images by calculating the Euclidean distance of the activation maps obtained from several layers of a pre-trained network, scaling them by weights $w$, and then averaging the values channel-wise to compute the final LPIPS score. 
\begin{figure*}[htb!]
    \centering
    \includegraphics[width=\textwidth]{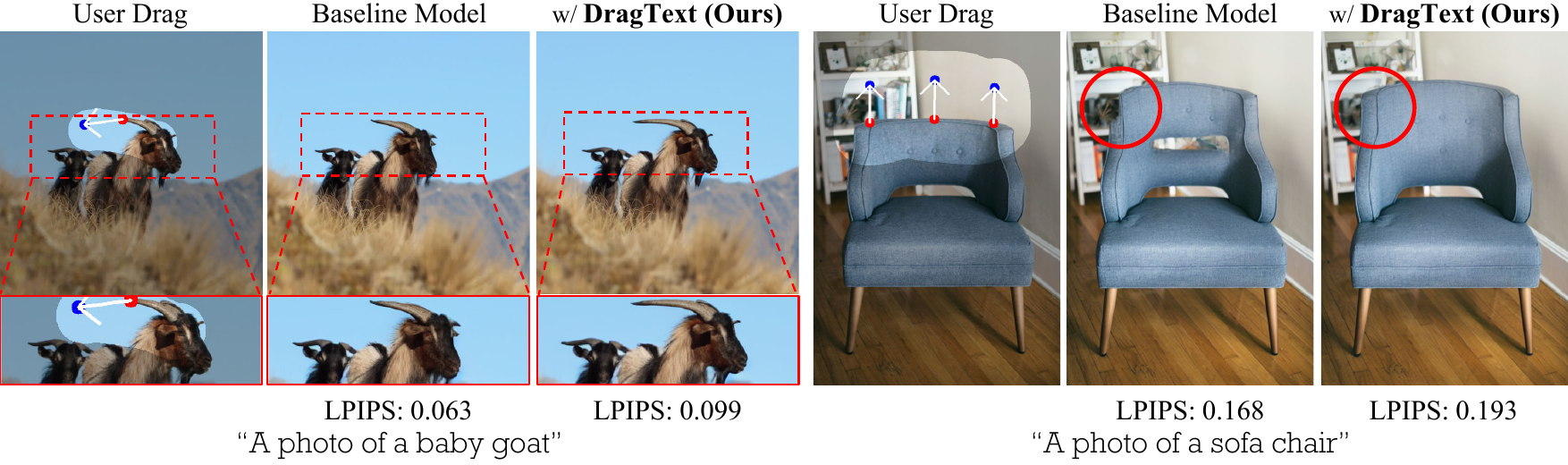}
    \vspace{-15pt}
    \caption{
        \textbf{Comparing the LPIPS scores of the baseline model and \textsc{DragText}.} 
        Despite \textsc{DragText} achieving better edits and preserving semantics well, its LPIPS scores are worse. 
        In the left image, the baby goat's face direction in the input image is the same in the result with \textsc{DragText}. 
        In contrast, the baseline model has the baby goat looking slightly more forward. 
        In the right image, unlike the baseline model, \textsc{DragText} maintains the complete form of the sofa chair.
    }
    \label{suppfig:lpips}
    \vspace{-10pt}
\end{figure*}

LPIPS is an appropriate metric for measuring the similarity between two images, emphasizing that image editing should maintain similarity to the original image. 
However, due to the nature of the drag editing task, the image will inevitably change. Consequently, even when dragging is performed successfully, the LPIPS score might worsen. 
For instance, if an image does not change at all, it would yield an LPIPS score of 0, the best possible score. 
As shown in \cref{suppfig:lpips}, even though we achieved a more desirable image editing outcome, the LPIPS score was lower. 
Therefore, we propose that LPIPS should not be overly emphasized if the score falls below a certain threshold. To address this issue, we suggest using the product of LPIPS and MD, which are complementary metrics, as a more robust evaluation metric.

\vspace{0.5cm}
\subsection{Mean Distance}
\begin{figure*}[htb!]
    \centering
    \includegraphics[width=\textwidth]{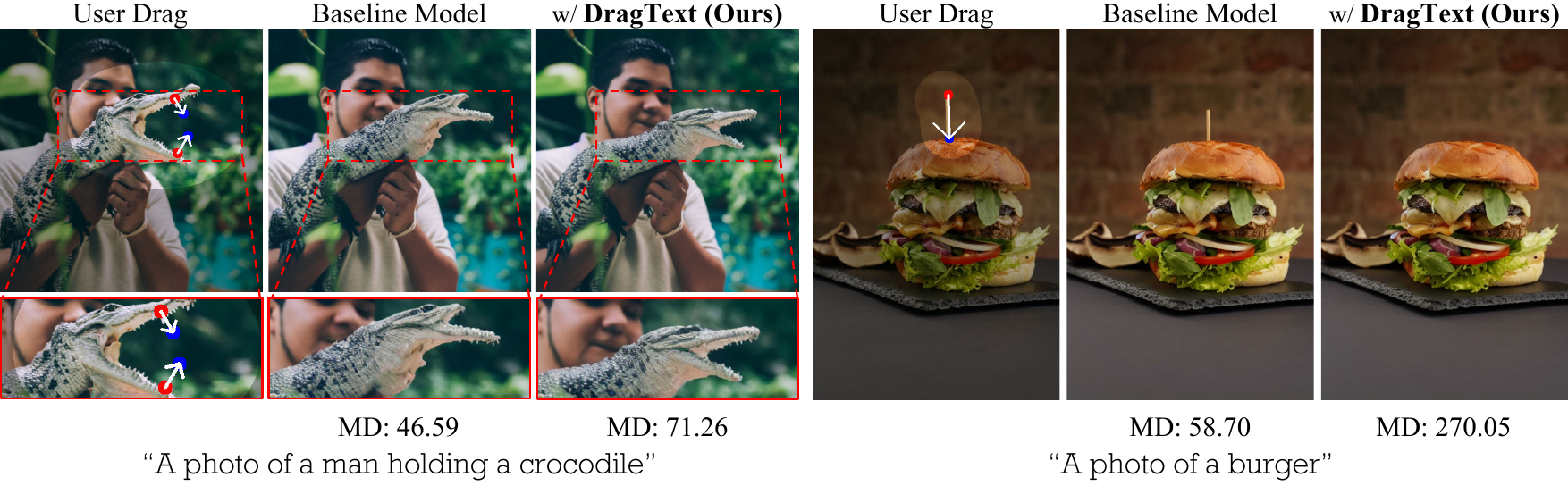}
    \vspace{-15pt}
    \caption{
        \textbf{Comparing the MD scores of the baseline model and \textsc{DragText}.} 
        Despite \textsc{DragText} moving the handle point closer to the target point, the MD score was lower.
        In the left image, the baseline model showed a better MD score, despite the crocodile's snout shape being distorted compared to \textsc{DragText}. 
        In the right image, \textsc{DragText} completed dragging, but its MD score is worse.
    }
    \label{suppfig:md}
    \vspace{-10pt}
\end{figure*}
Mean distance (MD) is computed via DIFT~\cite{DIFT}. 
First, DIFT identifies corresponding points in the edited image that correspond to the handle points in the original image. 
These identified points are regarded as the final handle points after editing is complete.
Then, the mean Euclidean distance between the corresponding point and the target point is calculated.
MD is the average value of all handle-target point pairs.

We propose that evaluating drag editing using Mean Distance (MD) on certain images in the DragBench dataset is challenging. 
Some images in DragBench require specific objects to disappear through drag editing as the points move. 
However, if a specific object disappears, there would be no corresponding objects in the edited image, resulting in a significantly high MD value. 
For instance, in \cref{suppfig:md}, the handle point and target point indicate that the toothpick should be perfectly inserted into the hamburger. 
Despite successfully achieving this, DIFT fails to recognize the toothpick, resulting in a higher MD value being calculated.
Conversely, there are cases where the MD value is low because the points remain in the same semantic position, but the actual image editing was unsuccessful due to distorted shapes and loss of semantics. 
While MD is an excellent metric for tasks involving moving feature points of objects, it has certain limitations and challenges when applied to all images in point-based editing tasks.

\vspace{1cm}
\section{Visual Ablation on the Hyperparameters of Regularization}\label{suppsec:regularization}
\begin{figure*}[htb!]
    \centering
    \includegraphics[width=\textwidth]{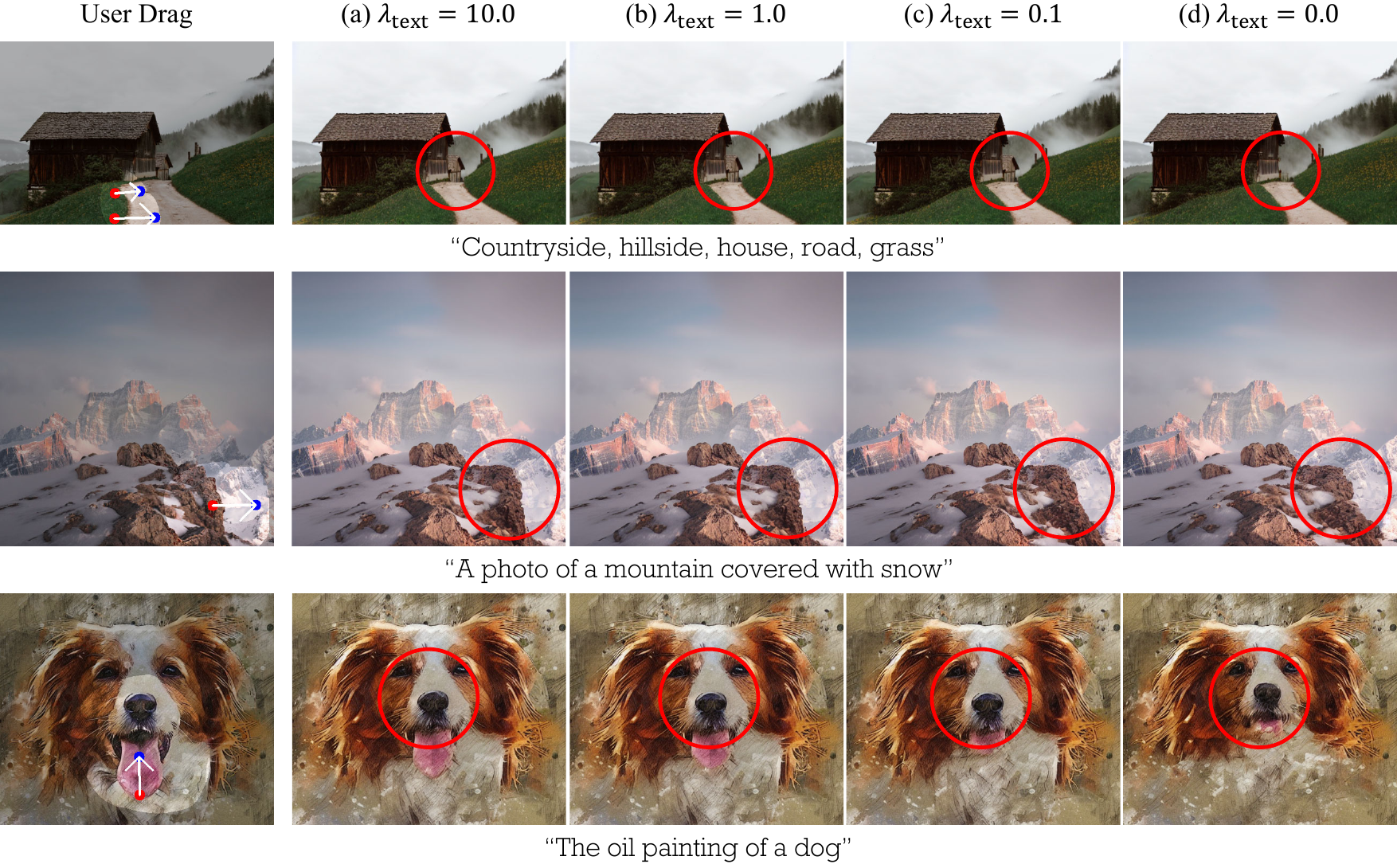}
    \vspace{-15pt}
    \caption{\textbf{Effect of text regularization with visual ablations on the hyperparameter $\lambda_\text{text}$.}
    The red circles indicate the loss of semantics. 
    In the first column, the small house on the right has disappeared. 
    In the second column, the cliff's shape has collapsed. 
    In the third column, the position of the dog's facial features has changed.
    }
    \label{suppfig:reg}
    \vspace{-10pt}
\end{figure*}

In \cref{suppfig:reg}, we provide extra visual ablation results to demonstrate how the hyperparameter $\lambda_\text{text}$ impacts the regularization process in text optimization. 
We modified images by adjusting $\lambda_\text{text}$ within a range from 0 to 10, which allowed us to control the level of regularization applied during the text optimization phase.
When $\lambda_\text{text}$ is close to 0, it results in some of the important semantic information being lost. 
On the other hand, applying an excessively high $\lambda_\text{text}$, prevents the optimization of the text embedding from effectively altering the image. 

\section{Ablation on the U-Net Feature Maps}\label{suppsec:unet}
\begin{figure*}[htb!]
    \centering
    \includegraphics[width=\textwidth]{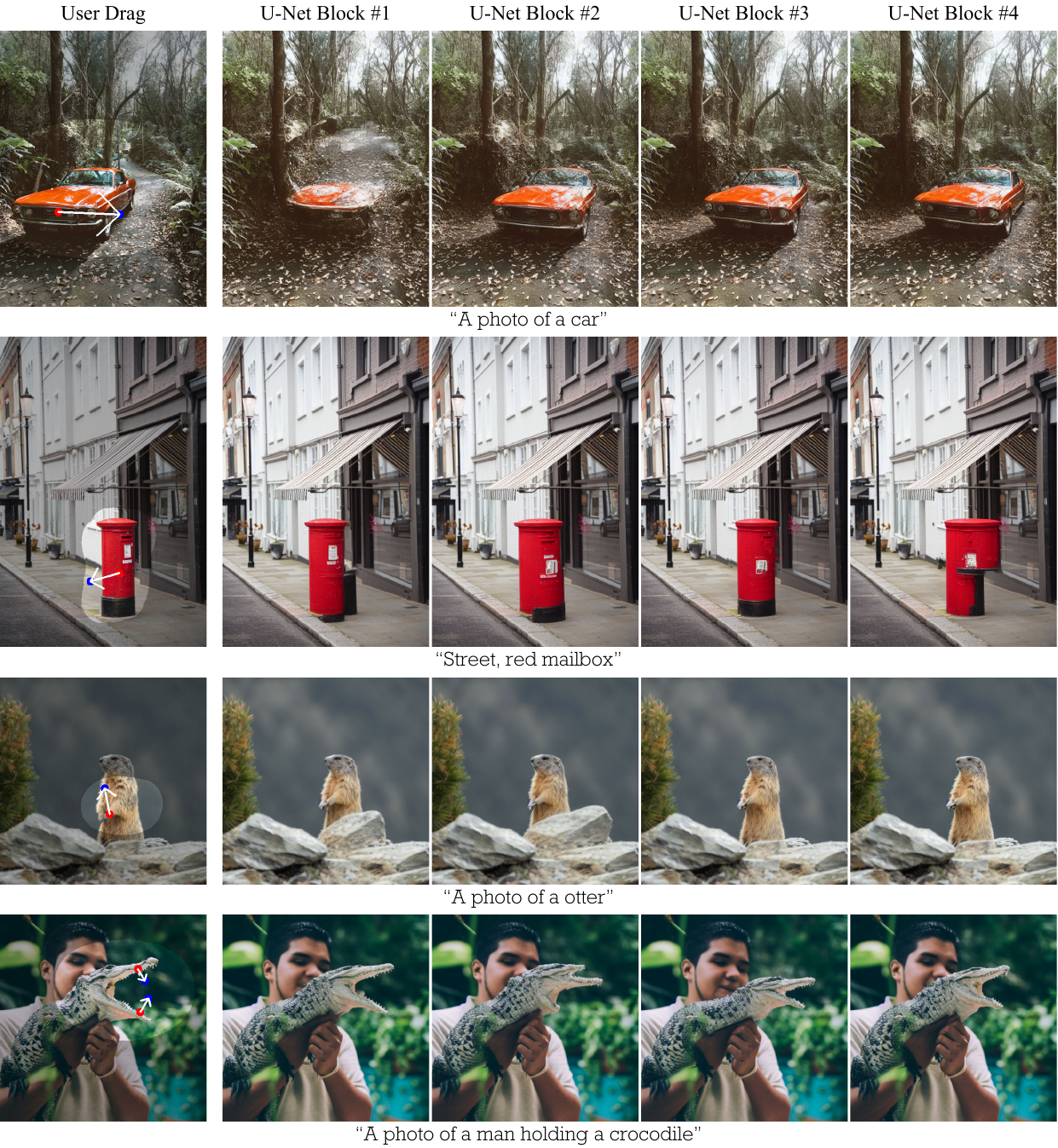}
    \vspace{-15pt}
    \caption{
    \textbf{More image generation results per U-Net decoder block.} Optimizing the text embedding using the 3rd block of the U-Net decoder yields the best performance in terms of dragging and image semantics.
    }
    \label{suppfig:unet}
    \vspace{-10pt}
\end{figure*}

We utilize various U-Net decoder blocks for \textsc{DragText} with the image embedding fixed from the 3rd block. 
In \cref{suppfig:unet} and \Cref{supptab:unet}, The 3rd block maintains semantics and achieves effective dragging. 
Lower blocks (\eg, Block 1) have difficulty with semantics, and higher blocks (\eg, Block 4) exhibit poor dragging.
\begin{table}[h]
    \centering
    \caption{Quantitative evaluation results by U-Net block number applied in \textsc{DragText}}
    \renewcommand{\arraystretch}{0.8} 
    \resizebox{0.35\columnwidth}{!}{%
    \begin{tabular}{c|cc}
    \toprule 
    U-Net Block \# & LPIPS$\downarrow$ & MD$\downarrow$ \\ \midrule\midrule
    Block \#1 & 0.135 & 36.02 \\
    Block \#2 & 0.154 & 37.08 \\
    Block \#3 & 0.124 & \textbf{31.96} \\
    Block \#4 & \textbf{0.119} & 33.60 \\ \midrule
    \end{tabular}}
    \label{supptab:unet}
\end{table}

\vspace{1cm}
\section{More Qualitative Results for Manipulating Embeddings}\label{suppsec:app}
\begin{figure*}[htb!]
    \centering
    \includegraphics[width=\textwidth]{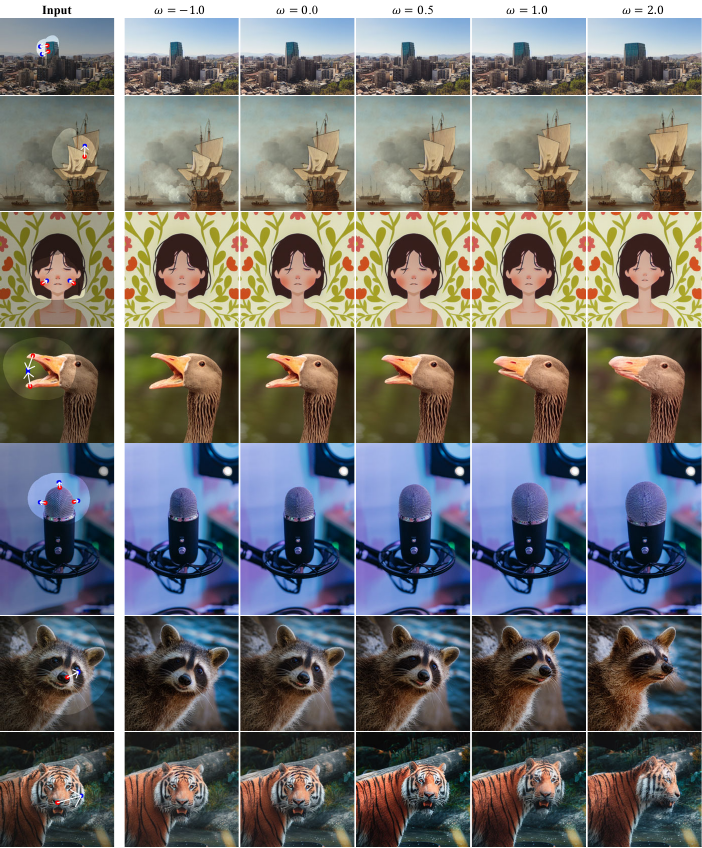}
    \vspace{-15pt}
    \caption{\textbf{More qualitative results for manipulating embeddings with \textsc{DragText}}.
    DragText can generate images outside the originally intended editing range in the direction of the drag while preserving the semantic content at the same time.
    }
    \label{suppfig:interp1}
    \vspace{-10pt}
\end{figure*}

\begin{figure*}[htb!]
    \centering
    \includegraphics[width=\textwidth]{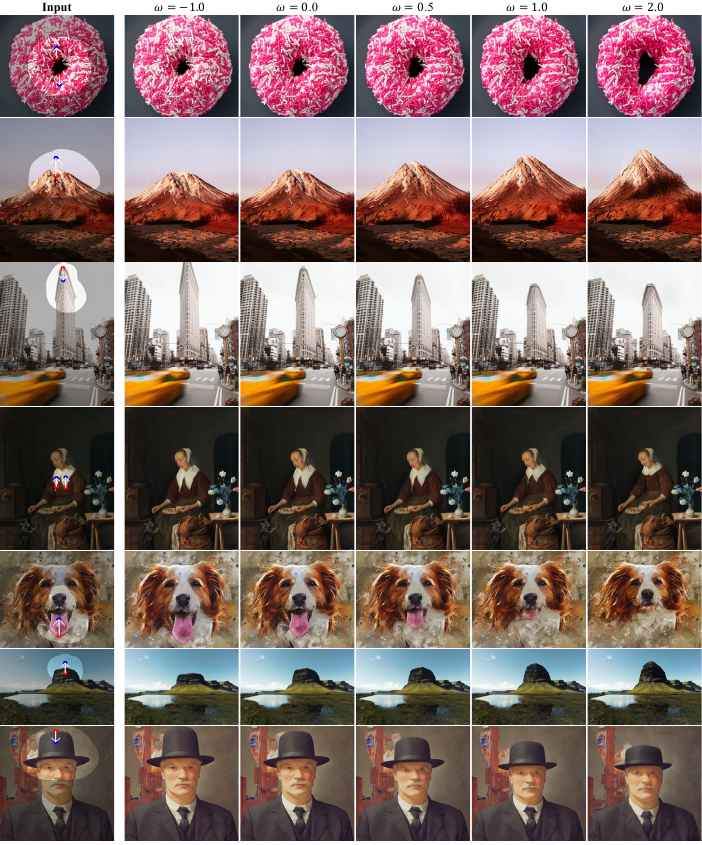}
    \vspace{-15pt}
    \caption{\textbf{More qualitative results for manipulating embeddings with \textsc{DragText}}.
    DragText can generate images outside the originally intended editing range in the direction of the drag while preserving the semantic content at the same time.
    }
    \label{suppfig:interp2}
    \vspace{-10pt}
\end{figure*}

In ~\cref{suppfig:interp1} and ~\cref{suppfig:interp2}, we apply linear interpolation and extrapolation to the image and text embeddings to generate not only the intermediate stages of the image editing process but also the parts beyond the editing process. 
This is possible because \textsc{DragText} optimizes both the text and image embeddings simultaneously.

\clearpage
\end{document}